\UseRawInputEncoding
\documentclass{article}

\usepackage{microtype}
\usepackage{graphicx}
\usepackage{subcaption}
\usepackage{booktabs} 
\usepackage{textcomp}
\usepackage{color}
\usepackage{xcolor}
\usepackage{url}
\usepackage{float}
\usepackage{longtable}
\usepackage{caption} 

\usepackage{hyperref}

\usepackage[preprint]{icml2026}

\usepackage{amsmath}
\usepackage{amssymb}
\usepackage{mathtools}
\usepackage{amsthm}

\numberwithin{figure}{section}
\numberwithin{table}{section}

\usepackage[capitalize,noabbrev]{cleveref}

\theoremstyle{plain}

\theoremstyle{definition}

\theoremstyle{remark}

\usepackage[textsize=tiny]{todonotes}

\icmltitlerunning{Evaluation of LLMs for Mathematical Formalization in Lean}

\begin{document}

\twocolumn[
  \icmltitle{Evaluation of LLMs for Mathematical Formalization in Lean}



  \icmlsetsymbol{equal}{*}

  \begin{icmlauthorlist}
    \icmlauthor{Tyson Klingner}{equal,yyy}
    \icmlauthor{Drew Bladek}{equal,yyy}
    \icmlauthor{Escher Crawford}{equal,yyy}
    \icmlauthor{Bohao Chen}{yyy}
    \icmlauthor{Ariel Fu}{yyy}
    \icmlauthor{Kaira Nair}{yyy}
    \icmlauthor{Jarod Alper}{yyy}
    \icmlauthor{Giovanni Inchiostro}{yyy}
    \icmlauthor{Vasily Ilin}{yyy}
  \end{icmlauthorlist}

  \icmlaffiliation{yyy}{Math AI Lab, University of Washington, Seattle, WA, USA}

  \icmlcorrespondingauthor{Tyson Klingner}{tysonk4@uw.edu}
  \icmlcorrespondingauthor{Drew Bladek}{abladek@uw.edu}
  \icmlcorrespondingauthor{Escher Crawford}{ecraw4d@uw.edu}
  \icmlcorrespondingauthor{Bohao Chen}{bohaoc@uw.edu}
  \icmlcorrespondingauthor{Ariel Fu}{xinyuf3@uw.edu}
  \icmlcorrespondingauthor{Kaira Nair}{kaira@uw.edu}
  \icmlcorrespondingauthor{Jarod Alper}{alper@uw.edu}
  \icmlcorrespondingauthor{Giovanni Inchiostro}{ginchios@uw.edu}
  \icmlcorrespondingauthor{Vasily Ilin}{vilin@uw.edu}

  \icmlkeywords{Machine Learning, ICML}

  \vskip 0.3in
]



\printAffiliationsAndNotice{}  

\begin{abstract}
    Within the past few years, the ability of Large Language Models (LLMs) to generate formal mathematical proofs has improved drastically. We provide a comparison of various LLMs' effectiveness in producing formal proofs in Lean 4 with the goal of assisting those seeking to use LLMs to support their own projects. We utilize both pass@$k$ and refine@$k$ metrics as the benchmark for our comparison and evaluate on subsets of both miniF2F and miniCTX datasets. Our testing shows that overall, Gemini 3.1 Pro and Claude Opus 4.7 perform best. Gemini 3.1 Pro achieved a 92\% success rate on miniF2F via refine@32 whereas Opus 4.7 achieved a 86\% success rate on miniCTX via refine@32. When taking cost into account, NVIDIA Nemotron 3 Super and GPT-OSS 120B were the most efficient, with competitive accuracies and average costs of $<\$0.01$ per correct proof.
\end{abstract}

\section{Introduction}

Proving mathematical theorems often requires a complex chain of arguments, logical deductions, and involves tedious computations. As such, the ability to prove theorems is emblematic of human intelligence and reasoning. Hence, evaluating a Large Language Model's (LLMs) ability to produce mathematical proofs provides an important benchmark to measure a given model's reasoning ability \cite{liu2025combibenchbenchmarkingllmcapability}. 
 A natural starting point in this direction is getting LLMs to produce natural language proofs \cite{zhang2026deeptheorem}. For example, advanced versions of Google's Deep Think achieved a gold-medal level performance in the 2025 International Mathematical Olympiad (IMO) \cite{DeepMindIMO2025}. However, it is currently difficult to verify the correctness of a natural language proof in an automated way \cite {lightman2024lets}. 
 
 Alternatively, we can task LLMs to produce a {\em formal language} proof \cite{zhou2025solvingformalmathproblems}. Formalization of mathematical proofs gives an objective metric to confirm a proof's correctness. Namely, we can use a formal theorem proving system such as Lean 4 \cite{10.1007/978-3-030-79876-5_37}.
 LLMs have recently shown strong improvements in formal proof generation in Lean 4 \cite{han2025simplifying, polu2023formal}. Most notably, Google's AlphaProof achieved a silver-medal performance in the 2024 IMO when producing formal proofs in Lean 4 \cite{1062014}.

 With an abundance of readily available LLMs it is natural to ask which LLM is the best at formal proof generation. The current options span from general-purpose frontier models, e.g. Google Gemini 3.1 Pro, to Lean specialized models, e.g. Goedel-Prover V2 \cite{lin2026goedelproverv} and Mistral's Leanstral \cite{mistral2026leanstral}. Moreover, for the general mathematician, there are many real world factors to consider before selecting a model for formal proof generation, including cost and access to computing power. In this paper, we compare several general and special purpose LLMs' capability in formal proof generation. More specifically, we fix an $n=50$ size subset of both miniF2F \cite{zheng2022miniff} and miniCTX \cite{hu2025minictx} datasets and utilize both the Pass@$k$ \cite{DBLP:journals/corr/abs-2107-03374} and Refine@$k$ \cite{xu2026icpceval} methodologies for $k=32$. See Table \ref{table of models} for a complete list of models considered. Beyond measuring proof success rates, we also analyze computational cost. Our goal is to provide a clear and practical comparison of different models and strategies, to help identify which approaches are most effective under real-world resource constraints.

 Our contributions are as follows:
{\setlength{\leftmargini}{2.0em}
 \begin{enumerate}
     \vspace{-0.15cm}\item \label{1} A comprehensive standardized evaluation of a wide array of LLMs for formal proof generation.
     \vspace{-0.15cm}\item \label{2} An extensive analysis of the cost efficiency.
     \vspace{-0.15cm}\item \label{3} A detailed comparison between the pass@$k$ and refine@$k$ methods.
     \vspace{-0.15cm}\item \label{4} A thorough categorization of error trends between models.
 \end{enumerate}
 }

 \section{Methodology}

 In this section, we explain every step and procedure in order so that the reader can reproduce our results. See \url{https://github.com/uw-math-ai/LLMsLean} for the explicit code used.
 \label{method}

 \subsection{Datasets}

 We construct our benchmark upon two complementary evaluation datasets: miniF2F \cite{zheng2022miniff}, which evaluates mathematical reasoning on isolated, competition-level problems (e.g., IMO, AMC, AIME), and miniCTX \cite{hu2025minictx}, which tests the model's ability to operate within a heavily-imported theorem structures intended to mimic real-world mathematical work.

 Evaluating frontier LLMs over the entirety of both datasets is expensive. To evaluate the models in a more cost-effective manner, we instead work over a subset of size $n=50$ for both miniF2F and miniCTX. To limit selection bias we employ sampling strategies based on the dataset’s structure. For miniF2F, we utilize proportional stratified sampling. We categorized problems into explicit strata (e.g., imo, amc12, aime, algebra, induction, numbertheory), and randomly sampled from each stratum in exact proportion to its frequency in the whole dataset. On the other hand, for miniCTX the problem domain is generally homogeneous and lacks explicit categorical distinctions, so we take a simple random sample across the total dataset. This approach ensures that both subsets remain statistically representative snapshots of their respective original benchmarks.

 \subsection{Models}

 We evaluate a comprehensive suite of models, structurally categorized by their purpose and availability: Proprietary Frontier Models, Open-Weight General-Purpose, and Lean-Specialized  Provers. Table \ref{table of models} lists the models considered and API endpoints utilized to ensure benchmark reproducibility. For Goedel, both models were run on NVIDIA H200 GPUs with 2 for 32B and 1 for 8B. We used 2 GPUs for 32B since initial testing showed 1 GPU had insufficient VRAM.

\begin{table*}[ht]
\centering
\caption{List of every general and special purpose LLM evaluated}
\label{table of models}
\begin{tabular}{lll}
\toprule
\textbf{Category} & \textbf{Model Family} & \textbf{Specific Target / API Identifier} \\ \midrule
\textbf{Proprietary Frontier} & \textbf{Google Gemini} & \begin{tabular}[c]{@{}l@{}}Gemini 3.1 Pro (gemini-3.1-pro-preview), \\ Gemini 3 Flash (gemini-3-flash-preview), \\ Gemini 3.1 Flash-Lite (gemini-3.1-flash-lite-preview)\end{tabular} \\ \cmidrule(lr){2-3}
\textbf{} & \textbf{Open AI GPT} & \begin{tabular}[c]{@{}l@{}}GPT-5.4 (gpt-5.4), GPT-5.4 Mini (gpt-5.4-mini), \\ GPT-5.4 Nano (gpt-5.4-nano)\end{tabular} \\ \cmidrule(lr){2-3}
 & \textbf{Anthropic Claude} & Claude Opus 4.7 (global.anthropic.claude-opus-4-7) \\ \midrule
\textbf{Open-Weight} & \textbf{Qwen} & Qwen 3.5 (Qwen3.5-397B-A17B) \\ \cmidrule(lr){2-3}
\textbf{General-Purpose} & \textbf{Nemotron} & Nemotron 3 Super (nemotron-3-super-120b-a12b) \\ \cmidrule(lr){2-3}
 & \textbf{Open AI OSS} & GPT-OSS-120B (gpt-oss-120b) \\ \cmidrule(lr){2-3}
 & \textbf{Deepseek} & DeepSeek 3.2 (DeepSeek-V3.2) \\ \midrule
\textbf{Lean Specialized} & \textbf{Goedel} & \begin{tabular}[c]{@{}l@{}}Goedel-Prover V2 (Goedel-Prover-V2-32B),\\ (Goedel-Prover-V2-8B)\end{tabular} \\ \cmidrule(lr){2-3}
 & \textbf{Mistral Labs} & Leanstral (labs-leanstral-2603) \\ \bottomrule
\end{tabular}
\end{table*}

 \subsection{Prompting and Model Settings}
 \label{section prompt}

 To ensure a fair comparison across a diverse field of LLMs, our framework uses a standardized zero-shot prompting paradigm. The agent is minimally contextualized and instructed to output the proof within a designated markdown block, assuming all Mathlib dependencies are pre-imported. We deliberately avoid model-specific prompt engineering or complex few-shot scaffolding, in order to evaluate the intrinsic formalization capability of the models. See Figures \ref{prompt1}, and \ref{prompt2} for the specific prompts.

To ensure that minor deviations from the requested formatting do not produce false negatives, our  extraction pipeline utilizes a cascading fallback strategy: it prioritizes the requested ‘FINAL’ templating, next checking for generic Lean tags (e.g., lean4), and ultimately locates theorem or lemma keywords if the model embeds the code in plain text.

We used a temperature of 0.5 in order to have enough variation in responses for the pass@$k$ metric to be useful, while maintaining enough consistency so that the refine@$k$ strategy would depend mostly on prompt feedback as opposed to randomness. See Section \ref{benchmarks} for the definition of pass@$k$ and refine@$k$. The exception to this is Opus 4.7, for which the API did not accept a temperature setting. The models were allowed a token budget of 16384 for reasoning, which was chosen to limit time and cost if the model got stuck in a loop while still allowing adequate tokens for thinking and output. 

 \subsection{Verification Protocols}

To prevent models bypassing the verification, our module sanitizes responses before compilation. Specifically, we filter out outputs containing ‘sorry’, ‘admit’, or ‘omitted’ macros, which are tactics that bypass proofs but satisfy the compiler. A generated proof is labeled as a {\em Pass} precisely when it compiles cleanly in the Lean 4 environment without errors or bypass statements, ensuring machine-checkable correctness.
 
Furthermore, to prevent models from altering theorem premises, we enforce strict structural extraction. Our pipeline isolates the proof body from the model's response and then combines it with the dataset’s formal statement. This structural safeguard guarantees that the Lean verifier evaluates the exact target theorem, neutralizing any hallucinated modifications to the problem definition. Evaluation was performed using LeanInteract 
to interface between our pipeline and the Lean Kernel.

 \subsection{Evaluation: Pass@$k$ and Refine@$k$}
 \label{benchmarks}

 We evaluate the models under two distinct generation methods, choosing a maximum of $k=32$ to balance computational costs with generating accurate and reliable data. Our methods are as follows:

\begin{itemize}
    \item {\bf Pass@$k$:} We generate $n=32$ independent proofs for each theorem at a temperature of $t=0.5$. The Pass@$k$ metric estimates the probability that at least one out of $k$ independent samples successfully verified in the Lean environment. \cite{DBLP:journals/corr/abs-2107-03374}. 
    Taking the average over the given dataset gives an unbiased expected value
    \begin{equation}
    \text{pass@}k := \underset{\text{Problems}}{\mathbb{E}} \left [ 1 - \frac{{{n-c}\choose{k}}}{{{n}\choose{k}}}\right ]
    \label{form}
    \end{equation}
    where $c$ is the number of correct samples out of the $n=32$ generations and $k$ is the number of solutions intended to be checked. See Figure \ref{passchart} for a flowchart visual.
    \item {\bf Refine@$k$:} To evaluate agentic formalization capabilities, we introduce an interactive refinement sequence akin to. If an initial generation fails, the exact trace from the Lean verifier is extracted. The model is re-prompted with its previous faulty proof alongside this specific compiler feedback, iterating for up to $k$ turns. 
    \begin{equation} 
    \text{Response}_i := 
    \begin{cases}
     LLM(\text{Prompt, FTS}) & i = 1 \\
     LLM(\text{Prompt}, \\ \text{FTS}, \text{Response}_{i-1}, \\ \text{Feedback}_{i-1}) & 1 < i \le k
    \end{cases}
    \end{equation}
    where $\text{FTS} := \text{Formal Theorem Statement}.$ We define refine@$k$ as the percentage of theorems that are correct after $k$ iterations, normalized to [0,1]. See Figure \ref{refinechart} for a flowchart visual.
\end{itemize}

\subsection{Model Costs}

To determine model pricing, we use two different methods. For models that have a dedicated API endpoint, we calculate the price based on the recorded input and output tokens consumed for each API call. For the models without a dedicated API endpoint we ran the experiments using a local HPC Cluster, which cost $\$0.90$/ GPU hr. However, to replicate the experiments the reader may have to pay market rate e.g.,  \cite{vastai_h200_pricing} which is approximately $\$3.48$/ GPU hr. All prices were obtained in May 2026, with model providers and prices indicated in Table \ref{cost of models}.


\begin{table}[H]
\caption{LLMs pricing \$USD / million tokens}
\label{cost of models}
\begin{tabular}{lcc}
\toprule
\textbf{Model} & \textbf{\begin{tabular}[c]{@{}l@{}}Input Cost\end{tabular}} & \textbf{\begin{tabular}[c]{@{}l@{}}Output Cost\end{tabular}} \\ \midrule
\textbf{Opus 4.7} & 5 & 25 \\ 
\textbf{Nemotron} & 0.3 & 0.9 \\ 
\textbf{Deepseek 3.2} & 0.3 & 0.45 \\ 
\textbf{Qwen 3.5} & 0.6 & 3.6 \\ 
\textbf{GPT 5.4} & 2.5 & 15 \\ 
\textbf{GPT 5.4-mini} & 0.75 & 4.5 \\ 
\textbf{GPT 5.4-nano} & 0.2 & 1.25 \\ 
\textbf{Gemini 3.1 Pro} & 2 & 12 \\ 
\textbf{Gemini 3 Flash} & 0.5 & 3 \\ 
\textbf{Gemini 3.1 Flash-Lite} & 0.25 & 1.5 \\ 
\textbf{GPT-OSS} & 0.15 & 0.6 \\ 
\textbf{Leanstral} & 0 & 0 \\ \bottomrule
\end{tabular}
\end{table}

\section{Results}

    \subsection{Accuracy}
    \label{accuracy}
    \begin{figure*}[h]
        \centering
        \subfloat[MiniCTX Accuracy vs. $k$ (Pass and Refine)]{%
            \includegraphics[width=1\linewidth]{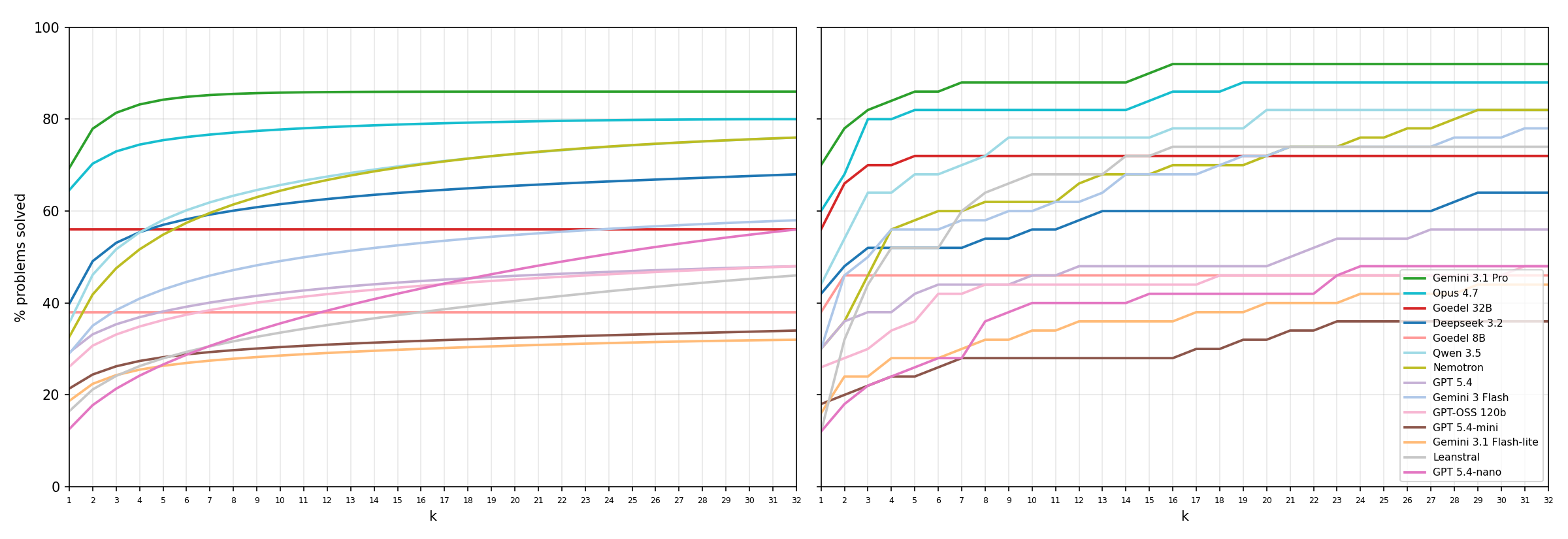}%
            \label{subfig:a}%
        }\hfill\\
        \subfloat[MiniF2F Accuracy vs. $k$ (Pass and Refine)]{
            \includegraphics[width=1\linewidth]{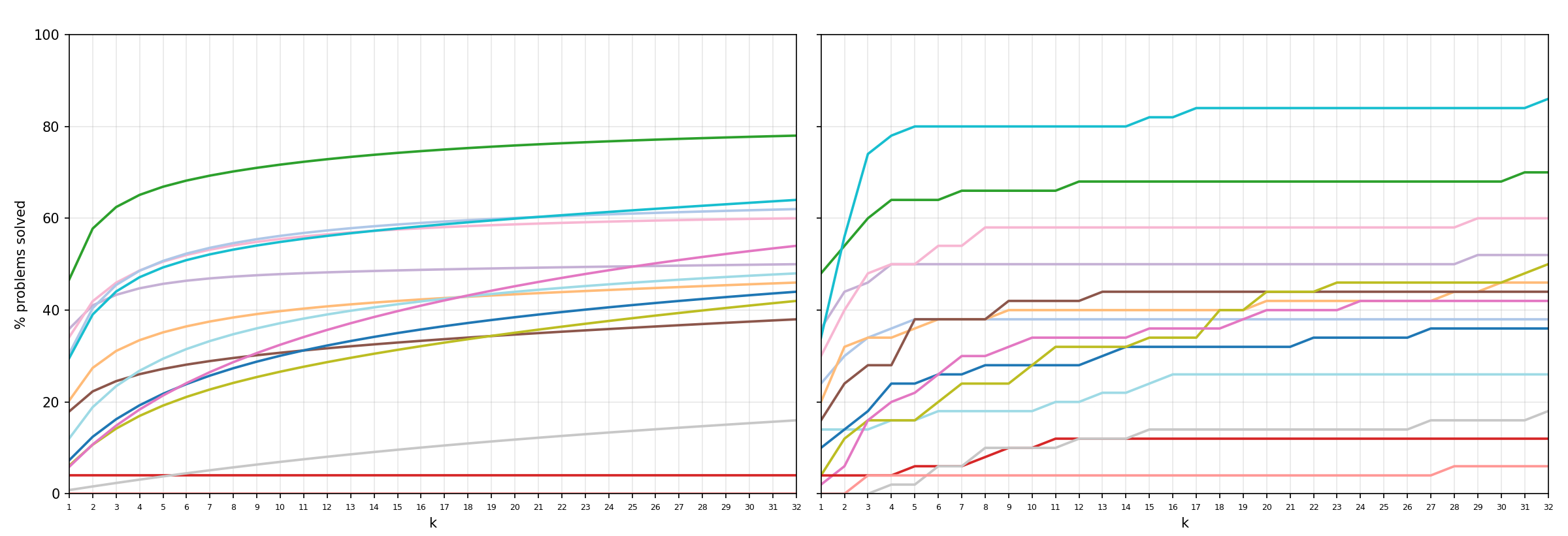}%
        }
        \caption{Each graph presents the pass@$k$/refine@$k$ for $1 \le k \le 32$ of all models on miniF2F and miniCTX. Each curve is increasing, with the exception of the Goedel models for pass@$k$, which did not increase. This is likely due to homogeneity in responses
        In general, accuracies on miniCTX were lower, with the exception of the frontier models (Gemini Pro etc.) and GPT-OSS. The curves for pass@$k$ are smoother since the pass@$k$ function is given by a binomial expression (\ref{form}), whereas  refine@$k$ has discrete interval values. For $k$ near 32, Gemini 3.1 Pro was the top performing model on all but refine@$k$ on miniCTX, where it is outperformed by Opus 4.7. On the other hand, Goedel 8B had the lowest pass@$k$ and refine@$k$ on miniCTX, Gemini 3.1 Flash-Lite had the lowest pass@$k$ on miniF2F, and GPT 5.4-mini had the lowest refine@$k$ on miniF2F.}
        \label{curvesbydataset}
    \end{figure*}

    Considering Figure \ref{curvesbydataset}, which presents the accuracy vs. $k$ curves for pass/refine@$k$ on each dataset and for all models evaluated. We see that Gemini 3.1 Pro and Claude Opus 4.7, both frontier models, were the two best-performing models for both pass@32 and refine@32 on both datasets. Nemotron and Qwen 3.5 performed well on miniF2F, achieving 3rd and 4th respectively, but struggled with miniCTX ($\sim20\%$ worse than the best model). On the other hand, GPT 5.4 and GPT-OSS were roughly average on miniF2F, but ranked right behind Gemini and Opus in miniCTX. 
    
    
    \par In addition, the specialized models (Leanstral, Goedel) performed reasonably well on miniF2F, with Goedel 32B and Leanstral performing above average via refine@32, as shown in Figure \ref{f2f32}. However, these models underperformed on miniCTX by a significant margin, trailing the next worst model by over 20\% on pass@32, with the Goedel models performing the worst\footnote{It is worth noting that the pass@$k$ lines for both Goedel models are horizontal, indicating a lack of diversity in responses, likely caused by our temperature of $t = 0.5$ not being properly applied.}, as indicated by Figure \ref{ctx32}. This may be due to overfitting for datasets like miniF2F that provide little to no context prior to the formal statement of the theorem needing proof. In order to challenge this strategy, the miniCTX dataset first formalizes new definitions and lemmas before giving a formal statement that depends on them \cite{hu2025minictx}. Thus, it requires a greater ability to generalize, as seen in models like Gemini and Opus. This also helps explain why general-purpose models have a smaller difference in accuracy between the two datasets.
    \par When considering the most practical model, it is also important to compare performances at low $k$, since many users may not want to waste time and money regenerating a proof dozens of times. From Figure \ref{curvesbydataset}, we can see that GPT-OSS performs almost as well as the top models on miniCTX, for $k \in \{1,2\}$. However, this is not the case for miniF2F, in which GPT-OSS performed below the median for low $k$.

    \subsection{Error Analysis}
    \label{error analysis sec}
     \begin{figure*}[]
        \centering
        \includegraphics[width=1\linewidth]{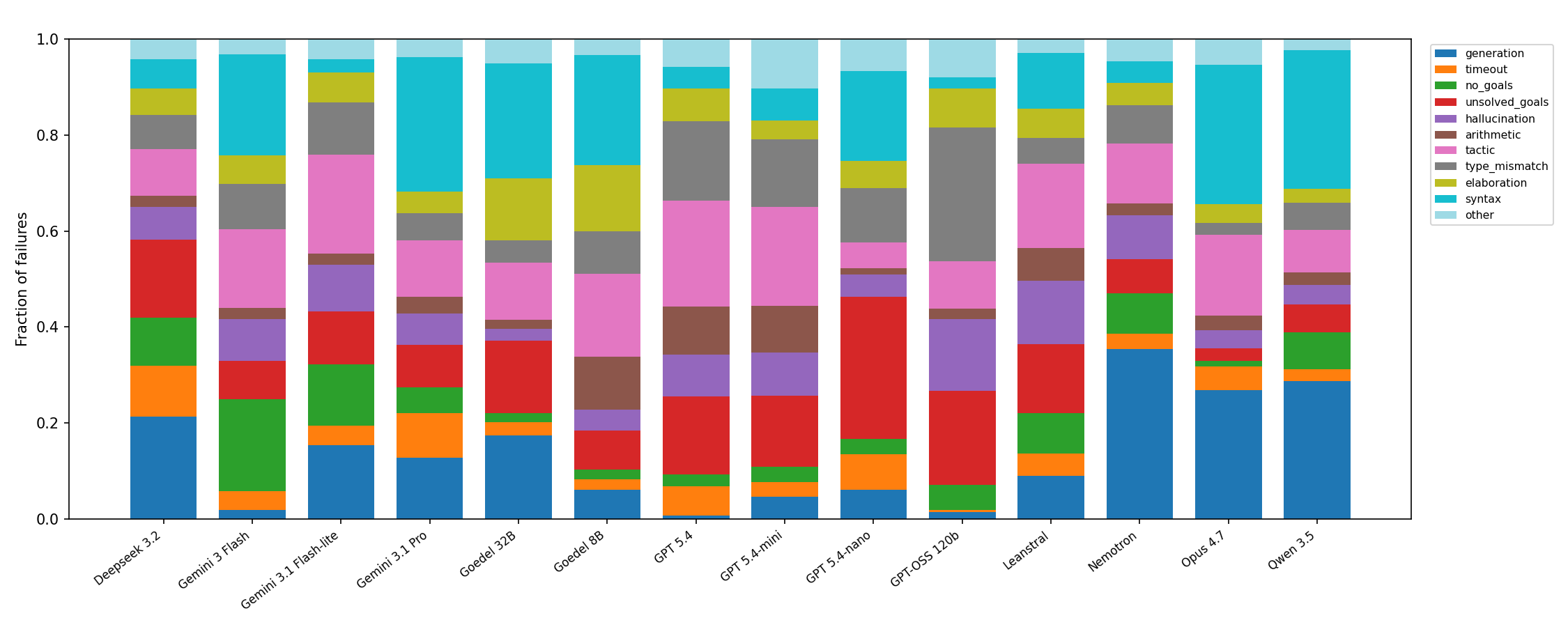}
        \caption{\textbf{Error category distribution per model}. See Appendix \ref{errortable} for a description of each error type and specific values. The top 4 categories (in order) are tactical errors (improper use of Lean tactics such as `\texttt{rw}', `\texttt{simp}', etc.), syntactical errors (incorrect Lean syntax), unsolved goals (incomplete proofs), and generation failures (use of `\texttt{sorry}'/`\texttt{admit}', failed API calls). Cumulatively, these 4 categories make up 54.7\% of errors.}
        \label{errorsbymodel}
    \end{figure*}
    First, we discuss generation errors of different models. One thing to note is that Nemotron, Qwen 3.5, and Deepseek 3.2 all have high proportions of generation failures, partially due to the API used returning an empty string when going over the token limit. However, most theorems take fewer than our $2^{14}$ token limit to solve, which indicates that these models are more prone to runaway responses. Opus also has a very high proportion of generation failures, mostly from the usage of `\texttt{sorry}' and `\texttt{admit}'. This indicates that Opus tends to admit defeat more readily than other models on difficult proofs. Taking into account Opus's overall high performance and its distribution of other error categories, it is also likely that Opus is very effective at mitigating strategical errors.

    \par Next, we analyze the prevalence of tactical errors throughout the models. GPT 5.4-nano has the smallest proportion of tactical errors, however, it also has the largest proportion of unsolved goals, implying that it tends to end proofs prematurely rather than make tactical mistakes. GPT 5.4 and Gemini 3 Flash also have a relatively large proportion of tactical errors while also having minimal generation errors. This means that these models rarely use `\texttt{sorry}', and prefer to make an attempt at solving the given problem no matter their confidence. Looking at Gemini 3 Flash in particular, it has the largest proportion of `no\_goals' errors, in which it continues to write tactics even after all goals have already been solved. This suggests that Gemini 3 Flash does not always know when it is correct.
    \par Regarding syntax errors, one interesting result is the disparity of  syntax errors between miniCTX and miniF2F. Per \ref{tab:minictxerror} and \ref{tab:minif2ferror}, we can see that syntax errors make up 10.7\% more of the errors on miniCTX as compared to miniF2F. This is likely because of the many lemmas and definitions introduced before each problem statement in miniCTX, which lends itself to a higher chance of improper syntax use. Hallucination errors are also more prevalent in responses on miniCTX, likely for similar reasons: models make up plausible sounding lemmas (such as `\texttt{Padic.valuation\_nonneg}' and `\texttt{Int.toNat\_add\_of\_nonneg}', both of which are not real Mathlib Lemmas) based off of the context that miniCTX provides.
    
    \subsection{Cost Efficiency}
    \label{cost efficiency}

     \begin{figure*}[h!]
        \subfloat[Accuracy vs. Cost for miniF2F pass@32]{%
            \includegraphics[width=.495\linewidth]{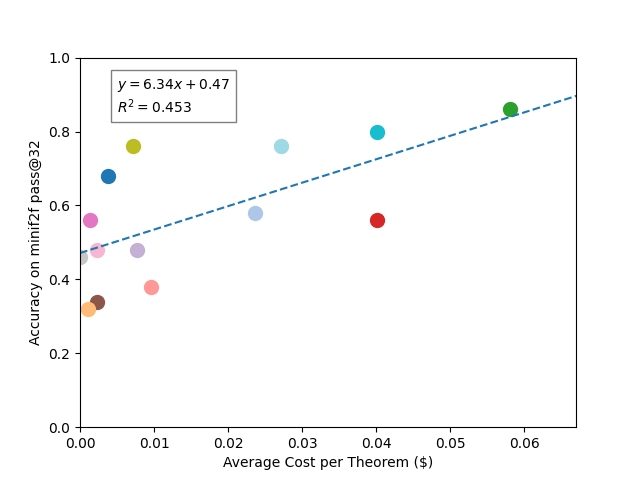}%
            \label{subfig:a}%
        }\hfill
        \subfloat[Accuracy vs. Cost for miniF2F refine@32]{%
            \includegraphics[width=.495\linewidth]{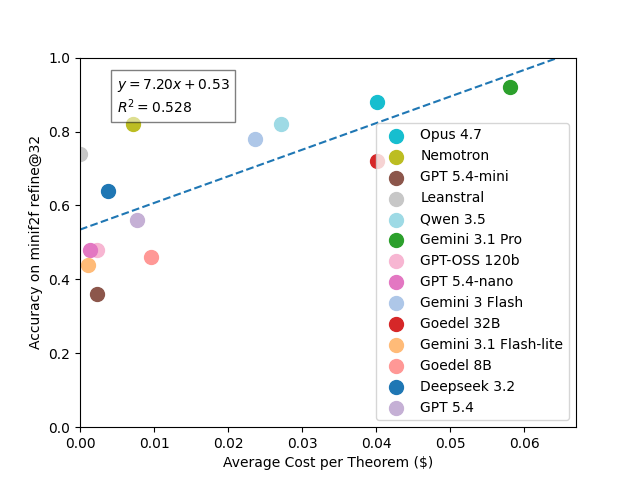}%
            \label{subfig:b}%
        }\\
        \subfloat[Accuracy vs. Cost for miniCTX pass@32]{%
            \includegraphics[width=.495\linewidth]{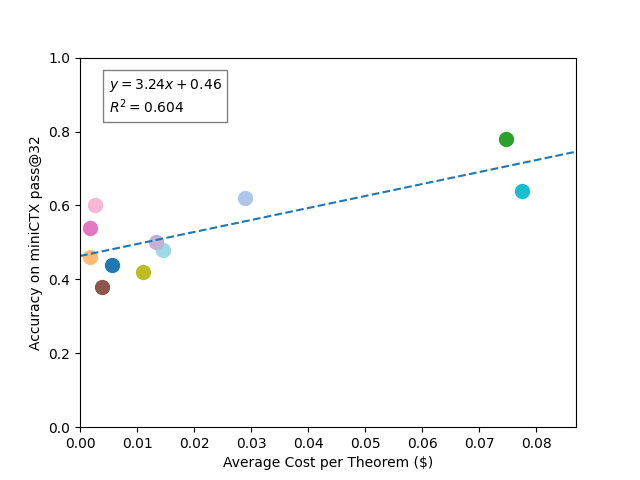}%
            \label{subfig:c}%
        }\hfill
        \subfloat[Accuracy vs. Cost for miniCTX refine@32]{%
            \includegraphics[width=.495\linewidth]{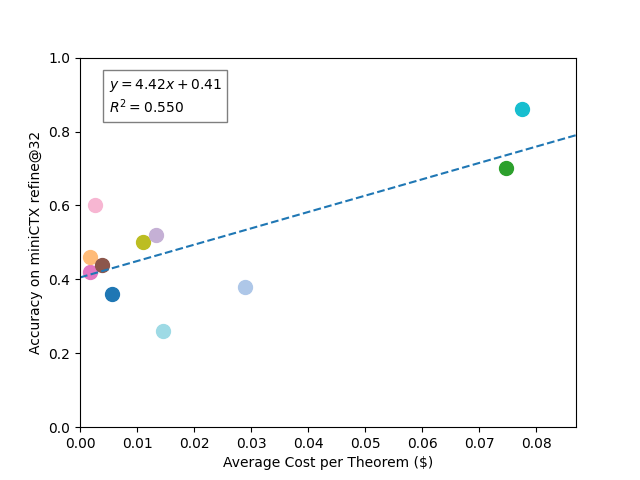}%
            \label{subfig:d}%
        }
        \caption{These plots show accuracy vs. average cost per attempt for miniF2F and miniCTX pass/refine@32. Accuracy scales linearly with average cost, with an increase of over $3\%$ per $\$0.01$ increase in cost per attempt. Most models are clustered below $\$0.03$ per attempt excluding most frontier models. Interestingly, GPT 5.4, which despite its frontier status used relatively few tokens. Refine@$k$ is also consistently marginally more expensive than pass@$k$, likely due to the increased input size.}
        \label{fig:fig}
    \end{figure*}

    When choosing a model, cost is often a major limiting factor. As shown in Figures \ref{subfig:a} and \ref{subfig:b}, model cost per theorem is correlated with improvements to effectiveness on miniF2F, with $R^2$s of 0.453 and 0.528. For miniCTX, the relation is much less strongly correlated, with $R^2$s of $0.108$ and $0.225$ (not shown). However, when we neglect the specialized models, which as discussed above did especially poorly on miniCTX, the $R^2$s increase to $0.604$ and $0.55$. Together, these trends suggest the emergent phenomenon of the formalization capability: larger, more expensive general-purpose models exhibit good generalization to formal proof generation, better than small specialized models. \\
    With that in mind, to determine the most cost-efficient model, we identified the degree of overperformance from expected by calculating the residuals of each plot and then ordering them. Table \ref{residuals} presents these residuals. The best performer for miniF2F was Nemotron, which outperformed expected by 24\% for pass@32 and 23\% for refine@32. For miniCTX however, Nemotron underperformed and GPT-OSS was the most effective, over-performing expected by 12\% for pass@32 and 18\% for refine@32. In addition, both these models cost around \$0.01 per attempt, making them ideal candidates for mathematicians on a budget. 

        
    
    \subsection{Refine Effectiveness}
    \label{refine effectiveness}
    It is important to analyze the per-model difference between the pass@$k$ and refine@$k$ metrics. For this, we define the difference metric $$\Delta_k := \text{refine@}k - \text{pass@}k.$$ Hence, a positive $\Delta_k$ means the model performed better with the iterative refine pipeline compared to the expected value of generating
    $k$ independent responses. On the other hand, a negative $\Delta_k$ says the opposite. 
    \begin{table}[h]
    \centering
    \caption{Model $\Delta_{32}$ on miniF2F and miniCTX benchmarks.}
    \label{tab:model_delta}
    \begin{tabular}{lrr}
    \toprule
    \textbf{Model} & \multicolumn{1}{c}{\textbf{miniF2F}} & \multicolumn{1}{c}{\textbf{miniCTX}} \\
    \midrule
    Opus 4.7      & $8.0$  & $\mathbf{22.0}$ \\ 
    Nemotron      & $6.0$  & $8.0$  \\ 
    Goedel 32B    & $16.0$ & $8.0$  \\ 
    Goedel 8B     & $8.0$  & $6.0$  \\ 
    GPT 5.4-mini  & $2.0$  & $6.0$  \\ 
    GPT 5.4       & $8.0$  & $2.0$  \\ 
    Leanstral     & $\mathbf{28.0}$ & $2.0$  \\ 
    Gemini Lite   & $12.0$ & $0.0$  \\ 
    GPT-OSS       & $0.0$  & $0.0$  \\ 
    Deepseek      & $-4.0$  & $-8.0$  \\ 
    Gemini Pro    & $6.0$  & $-8.0$  \\ 
    GPT 5.4-nano  & $-8.0$  & $-12.0$ \\ 
    Qwen 3.5      & $6.0$  & $-22.0$ \\ 
    Gemini Flash  & $20.0$ & $-24.0$ \\ 
    \midrule
    \textbf{Average} & $7.7$ & $-1.4$ \\
    \bottomrule
    \end{tabular}
    \end{table}
    \par Table \ref{tab:model_delta} presents the $\Delta_{32}$ values of all models on both miniF2F and miniCTX, sorted from greatest to least by miniCTX $\Delta_{32}$. On the miniF2F data subset, all but two of the models (Deepseek and GPT 5.4-nano) have a positive $\Delta_{32}$, with average  $\overline\Delta_{32}$ =  +7.7\%. However, on the miniCTX data subset, only 9 of the 14 models had a positive $\Delta_{32}$, with the average being $\overline\Delta_{32}$ = -1.4\%. Despite this, the models performed better on average with the iterative refinement method, with an overall average improvement of +3.14\%.
    \par On the miniCTX dataset, most of the models that performed better on refine@32 are frontier models (Gemini, GPT, Claude) that have very large context windows. We suspect this to be the primary limitation of this strategy---smaller models that lack the ability to store enough context begin to lose information on errors from previous attempts, especially due to the verbosity of some Lean 4 error messages, and thus are less likely to remedy the error in their next response. This issue would be further exacerbated by the complexity of the problems in the miniCTX dataset, many of which require longer formal problem statements that define theorems and lemmas necessary to solve the problem. This disparity in complexity between miniF2F and miniCTX likely explains why the iterative refinement strategy was less effective on the latter.

\section{Conclusion}

\subsection{Summary}

We evaluated a multitude of models on unbiased size $n=50$ subsets of both the miniCTX and miniF2F datasets, where we compared the overall effectiveness, cost-effectiveness and iterative refinement capabilities of each model. Based on these evaluations, we have identified Gemini 3.1 Pro and Claude Opus 4.7 as the current best models for formal proof generation in Lean 4. However, these models incurred a significant cost, up to an average of \$0.075 per attempt. In contrast, we have identified Nemotron and GPT-OSS as budget-friendly alternatives that still provide significant theorem proving capabilities, while reducing costs by up to 96\%. Across all models, iterative refinement consistently outperforms repeated independent attempts, indicating that feedback-driven reasoning loops are a key factor in improving formal proof generation.

\subsection{Limitations}
\label{limitations}

\subsubsection{Limited Scaling}

Our evaluations are limited to pass@32 and refine@32 benchmarks. In previous works, authors have evaluated models via pass@$k$ on various datasets with high $k$. For example, Goedel-Prover V2 has been tested on the PutnamBench up to pass@184 \cite{lin2026goedelproverv}, and Kimina-Prover has been tested on miniF2F up to pass@8192 \cite{wang2025kiminaproverpreviewlargeformal}. While higher $k$ can capture the model’s full potential for auto-formalization and increase accuracy for lower k measurements, it risks obfuscating the actual capabilities of a model when in use. Importantly, lower $k$ values better reflect real-world usage, where repeatedly sampling hundreds of outputs is often impractical, especially for high-cost models.  However, a larger amount of samples would increase confidence for $k=32$ and improve accuracy of our measurements. 

\subsubsection{Data Subsets}

Though we have taken care to make sure the subsets we used were as close to representative of the original datasets as possible, it is possible that there were some nuances, such as particularly difficult or easy problems that were over- or under-included.

\subsubsection{Standardized Prompt}

The standardized prompt may misrepresent some models' true accuracy. Indeed, prompt engineering has been shown to vastly improve results in other benchmarks, and the ideal prompt varies from model to model \cite{he2024doespromptformattingimpact}. For example, both Goedel-Prover V2  8B and 32B are designed on a specialized prompt, and achieved 84.6\% and 88.1\% on miniF2F via pass@32 respectively \cite{lin2026goedelproverv}. However, using a standardized prompt normalizes the results. Further, it better represents a model's capabilities when integrated into an agentic workflow, where prompts are nonstandard and can include large amounts of tooling specific to the agentic tool \cite{zhou2026multiagentdesignoptimizingagents}.

\subsubsection{Refine@$k$ is not Statistical}
\label{not statistical}

While pass@$k$ provides an expected value for the accuracy of each model when prompted $k$ times, refine@$k$ only provides a snapshot of what is possible with an iterative refinement strategy. However, when considering the miniF2F dataset, given the consistent improvement most models showed over Pass@$k$, it is likely that an iterative refinement strategy does in general improve results over repeated single attempt generation. This is consistent with previous work evaluating LLMs' reasoning ability with competitive programming contests via both pass@$k$ and refine@$k$ benchmarks \cite{xu2026icpceval}. On the other hand, when treating the miniCTX dataset, refine@$k$ underperformed pass@$k$ on GPT 5.4-nano, Gemini 3.1 Pro, and Qwen 3.5, see Figure \ref{ctx32}. We were not able to re-perform the experiments for financial reasons and this observation reveals the need for a statistical refine@$k$.


\subsection{Future Work}

The natural extension of our work is evaluating more LLMs. While we covered many well-known models, budgetary constraints limited our capabilities, and we were not able to test every model we initially considered. Additionally, as newer models are released, capabilities will change. Hence,  a website that compiles all the data and regularly updates similar to \url{https://llm-stats.com/} would be a worthy endeavor. Furthermore, as models improve, our current datasets will become saturated, even with a lower $k$ value. Thus, extending this work to other datasets could yield new insights on how well LLMs handle actual mathematical work. One candidate dataset for evaluation is RLMEval \cite{poiroux-etal-2025-rlmeval}, which focuses on formal research-level mathematical problems collected from a variety of mathematical topics. Another direction could be evaluating refine@$k$ in a statistical method. One method to accomplish this could be to choose a lower $k$, such as $k=4$ or $k=8$, and evaluate that $n$ times as in pass@$k$. This would provide a statistical look at refine@$k$ while keeping inference costs manageable.

\section*{Acknowledgments} 

CPU and GPU computing, LLM inference and data storage were in part done using AWS credits from the UW eScience School and UW IT. 
GPU computing was in part done on UWIT's GPU cluster Tillicum. 
GPU computing and model inference were in part done using TokenFactory.


\bibliography{ref}
\bibliographystyle{icml2026}

\newpage
\appendix
\onecolumn
\section{Additional Materials}
\subsection{Additional Information on Error Classifications}
\label{errortable}
    \begin{table}[H]
    \centering
    \caption{Error categories and their descriptions.}
    \renewcommand{\arraystretch}{1.4}
    \begin{tabular}{@{} l p{10cm} @{}}
    \toprule
    \textbf{Category} & \textbf{Description} \\
    \midrule
    \texttt{generation}     & Model produced a non-runnable proof: used \texttt{sorry}/\texttt{admit}, returned an empty string, or the API call failed. \\
    \texttt{timeout}        & The Lean server exceeded its heartbeat limit during verification. \\
    \texttt{syntax}         & Output is not parseable Lean (unexpected tokens, \texttt{import} statements, unterminated blocks). \\
    \texttt{unsolved\_goals} & Proof terminates with open goals remaining, including incomplete case splits. \\
    \texttt{no\_goals}       & Proof continues after all goals have been closed. \\
    \texttt{tactic}         & A specific tactic failed: \texttt{rewrite}/\texttt{rw} found no matching pattern, \texttt{simp}/\texttt{simp\_all} made no progress, \texttt{introN}/\texttt{rcases}/\texttt{ext} failed, or similar. \\
    \texttt{arithmetic}     & A decision procedure (\texttt{omega}, \texttt{linarith}, \texttt{norm\_num}, \texttt{ring}) could not close the goal. \\
    \texttt{type\_mismatch}  & Explicit type mismatch, bad anonymous constructor, or application type mismatch. \\
    \texttt{elaboration}    & Typeclass synthesis failure, and other errors that occur in the Lean 4 elaboration phase \\
    \texttt{hallucination}  & Unknown identifier, constant, or tactic --- the model referenced a name that does not exist in scope. \\
    \texttt{other}          & Uncategorized residual (5.0\%). \\
    \bottomrule
    \end{tabular}
    \label{tab:error-categories}
    \end{table}

    \begin{table}[ht]
    \caption{Per-model error profile (fraction of that model's failure entries).}
    \centering
    \tiny
    \renewcommand{\arraystretch}{1.3}
    \label{tab:per-model-error}
    \begin{tabular}{@{} l r r r r r r r r r r r @{}}
    \toprule
    Model & gen. & timeout & no\_goals & unsolved & halluc. & arith. & tactic & type & elab. & syntax & other \\
    \midrule
    Nemotron               & 0.354 & 0.032 & 0.084 & 0.072 & 0.091 & 0.025 & 0.124 & 0.081 & 0.046 & 0.045 & 0.047 \\
    Qwen 3.5               & 0.287 & 0.025 & 0.078 & 0.058 & 0.041 & 0.026 & 0.089 & 0.057 & 0.029 & 0.289 & 0.024 \\
    Opus 4.7               & 0.269 & 0.049 & 0.011 & 0.026 & 0.038 & 0.031 & 0.168 & 0.025 & 0.040 & 0.289 & 0.054 \\
    Deepseek 3.2           & 0.213 & 0.106 & 0.099 & 0.163 & 0.069 & 0.024 & 0.097 & 0.070 & 0.056 & 0.061 & 0.041 \\
    Goedel 32B             & 0.174 & 0.027 & 0.019 & 0.152 & 0.025 & 0.018 & 0.120 & 0.045 & 0.130 & 0.238 & 0.051 \\
    Gemini 3.1 Flash-Lite  & 0.153 & 0.041 & 0.129 & 0.109 & 0.098 & 0.023 & 0.205 & 0.110 & 0.061 & 0.028 & 0.042 \\
    Gemini 3.1 Pro         & 0.127 & 0.093 & 0.053 & 0.090 & 0.065 & 0.035 & 0.117 & 0.057 & 0.045 & 0.281 & 0.037 \\
    GPT 5.4-nano           & 0.060 & 0.075 & 0.032 & 0.295 & 0.047 & 0.013 & 0.054 & 0.112 & 0.057 & 0.187 & 0.066 \\
    Leanstral              & 0.089 & 0.047 & 0.084 & 0.144 & 0.132 & 0.068 & 0.175 & 0.054 & 0.061 & 0.117 & 0.029 \\
    GPT-OSS           & 0.014 & 0.005 & 0.052 & 0.195 & 0.151 & 0.022 & 0.098 & 0.278 & 0.082 & 0.023 & 0.080 \\
    GPT 5.4                & 0.007 & 0.061 & 0.025 & 0.162 & 0.087 & 0.101 & 0.220 & 0.166 & 0.068 & 0.046 & 0.057 \\
    Gemini 3 Flash         & 0.018 & 0.040 & 0.191 & 0.080 & 0.087 & 0.023 & 0.164 & 0.095 & 0.060 & 0.210 & 0.032 \\
    GPT 5.4-mini           & 0.046 & 0.030 & 0.032 & 0.148 & 0.090 & 0.098 & 0.205 & 0.141 & 0.040 & 0.066 & 0.103 \\
    Goedel 8B              & 0.061 & 0.022 & 0.020 & 0.081 & 0.044 & 0.110 & 0.174 & 0.088 & 0.138 & 0.230 & 0.033 \\
    \bottomrule
    \end{tabular}
    \end{table}

    \begin{table}[]
    \centering
    \renewcommand{\arraystretch}{1.3}
    \caption{Overall error category distribution.}
    \label{tab:error-distribution}
    \begin{tabular}{@{} l r r @{}}
    \toprule
    Category & Count & \% of failures \\
    \midrule
    \texttt{tactic}          & 8{,}128 & 14.4 \\
    \texttt{syntax}          & 7{,}987 & 14.2 \\
    \texttt{unsolved\_goals} & 7{,}519 & 13.3 \\
    \texttt{generation}      & 7{,}243 & 12.8 \\
    \texttt{type\_mismatch}  & 5{,}706 & 10.1 \\
    \texttt{hallucination}   & 4{,}366 &  7.7 \\
    \texttt{elaboration}     & 3{,}807 &  6.7 \\
    \texttt{no\_goals}       & 3{,}727 &  6.6 \\
    \texttt{arithmetic}      & 2{,}592 &  4.6 \\
    \texttt{timeout}         & 2{,}540 &  4.5 \\
    \texttt{other}           & 2{,}828 &  5.0 \\
    \bottomrule
    \end{tabular}
    \end{table}

    \begin{table}[]
    \centering
    \renewcommand{\arraystretch}{1.3}
    \begin{minipage}{0.47\textwidth}
    \centering
    \caption{miniCTX error distribution (total fails: 32,559).}
    \label{tab:minictxerror}
    \begin{tabular}{@{} l r r @{}}
    \toprule
    Category & Count & \% \\
    \midrule
    \texttt{generation}      & 3{,}325 & 10.2 \\
    \texttt{timeout}         & 2{,}021 &  6.2 \\
    \texttt{no\_goals}       & 1{,}424 &  4.4 \\
    \texttt{unsolved\_goals} & 3{,}207 &  9.8 \\
    \texttt{hallucination}   & 3{,}017 &  9.3 \\
    \texttt{arithmetic}      &   190 &  0.6 \\
    \texttt{tactic}          & 5{,}381 & 16.5 \\
    \texttt{type\_mismatch}  & 2{,}974 &  9.1 \\
    \texttt{elaboration}     & 3{,}131 &  9.6 \\
    \texttt{syntax}          & 6{,}074 & 18.7 \\
    \texttt{other}           & 1{,}815 &  5.6 \\
    \bottomrule
    \end{tabular}
    \end{minipage}
    \hfill
    \begin{minipage}{0.47\textwidth}
    \centering
    \caption{miniF2F error distribution (total fails: 23,884).}
    \label{tab:minif2ferror}
    \begin{tabular}{@{} l r r @{}}
    \toprule
    Category & Count & \% \\
    \midrule
    \texttt{generation}      & 3{,}918 & 16.4 \\
    \texttt{timeout}         &   519 &  2.2 \\
    \texttt{no\_goals}       & 2{,}303 &  9.6 \\
    \texttt{unsolved\_goals} & 4{,}312 & 18.1 \\
    \texttt{hallucination}   & 1{,}349 &  5.6 \\
    \texttt{arithmetic}      & 2{,}402 & 10.1 \\
    \texttt{tactic}          & 2{,}747 & 11.5 \\
    \texttt{type\_mismatch}  & 2{,}732 & 11.4 \\
    \texttt{elaboration}     &   676 &  2.8 \\
    \texttt{syntax}          & 1{,}913 &  8.0 \\
    \texttt{other}           & 1{,}013 &  4.2 \\
    \bottomrule
    \end{tabular}
    \end{minipage}
    \end{table}
    
    \subsection{Additional Accuracy Information}
        \begin{figure}[]
    \centering
    \begin{minipage}{0.45\textwidth}
        \centering
        \includegraphics[width=1\textwidth]{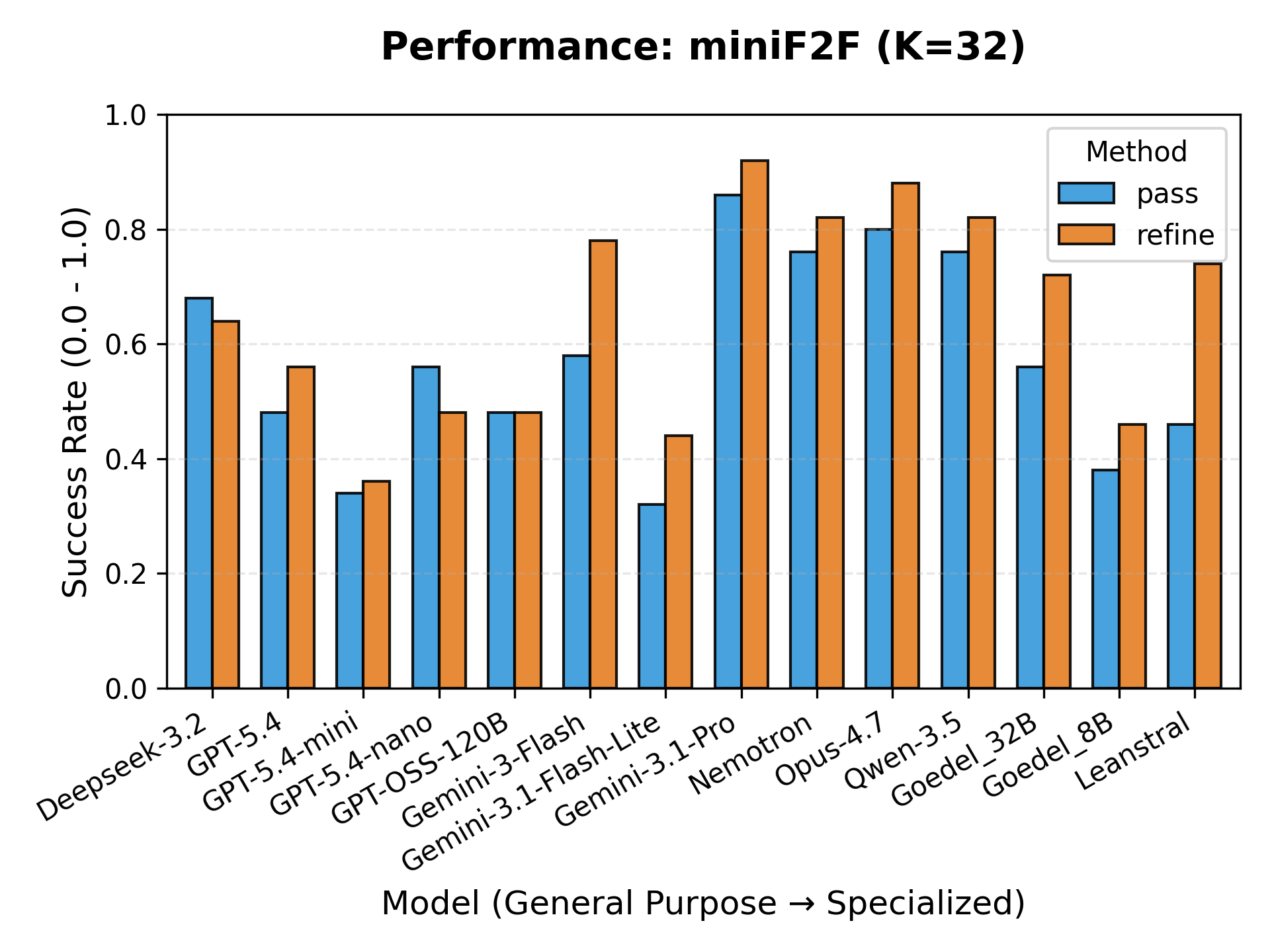} 
        \caption{Model pass@32 and refine@32 on miniF2F}
        \label{f2f32}
    \end{minipage}
    \begin{minipage}{0.45\textwidth}
        \centering
        \includegraphics[width=1\textwidth]{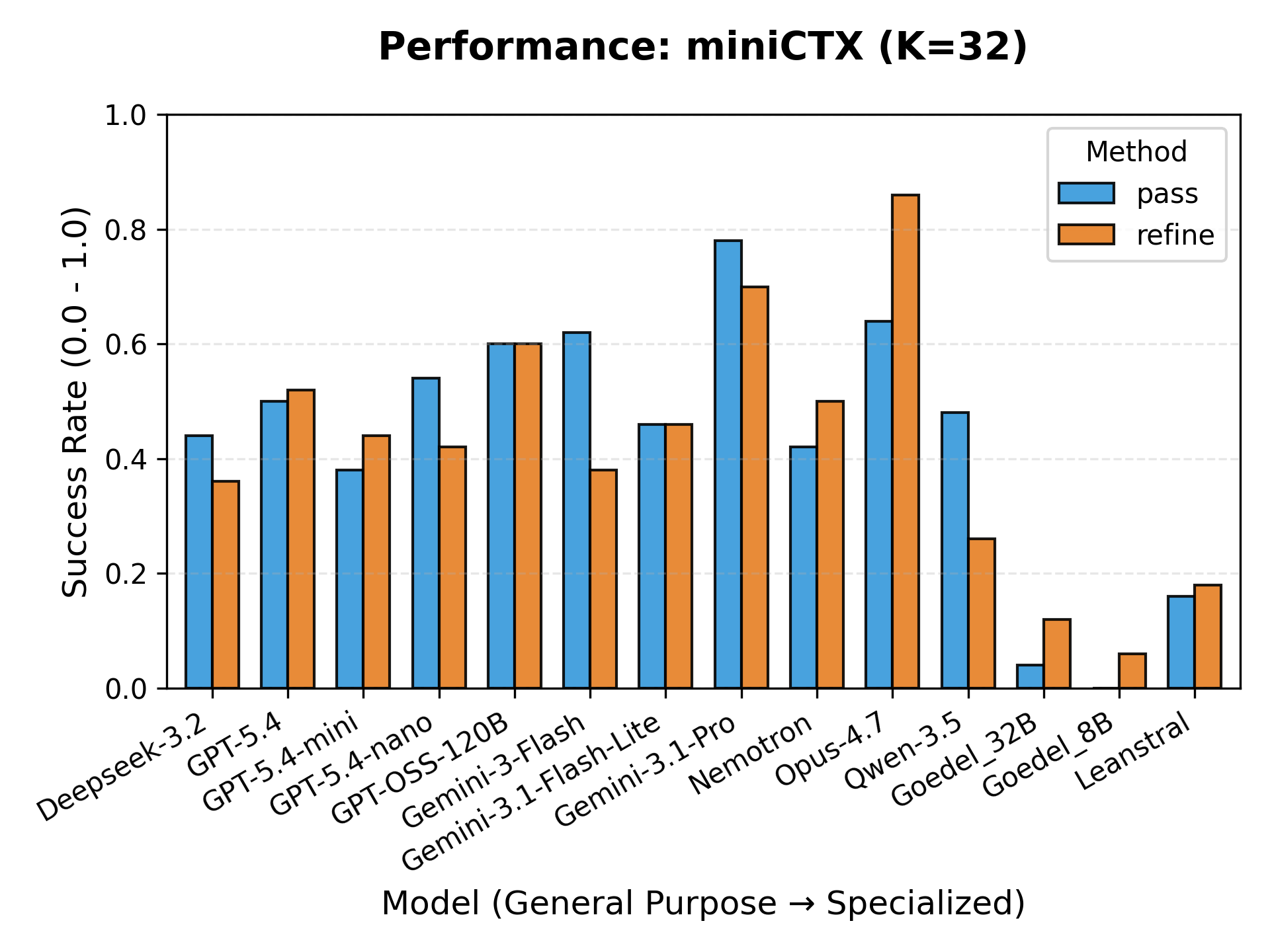} 
        \caption{Model pass@32 and refine@32 on miniCTX}
        \label{ctx32}
    \end{minipage}
    \end{figure}

    \clearpage

    \subsection{Additional Cost Information}

    \begin{center}
    \captionof{table}{Model Cost Residuals}
    \label{residuals}
    \begin{tabular}{lcccc}
        \toprule
        \textbf{Model} & \textbf{F2F Pass@32} & \textbf{F2F Refine@32} & \textbf{CTX Pass@32} & \textbf{CTX Refine@32} \\ \midrule
        Nemotron & \textbf{0.24} & \textbf{0.23} & -0.08 & 0.05 \\ 
        Deepseek 3.2 & 0.18 & 0.08 & -0.04 & -0.07 \\ 
        Qwen 3.5 & 0.12 & 0.09 & -0.03 & -0.21 \\ 
        GPT 5.4-nano & 0.08 & -0.06 & 0.07 & 0.01 \\ 
        Opus 4.7 & 0.07& 0.06 & -0.07 & 0.11 \\ 
        Gemini 3.1 Pro & 0.02 & -0.03 & 0.07 & -0.04 \\ 
        GPT-OSS & -0.01 & -0.07 & \textbf{0.13} & \textbf{0.18} \\ 
        Leanstral & -0.01 & 0.21 & N/A & N/A \\ 
        GPT 5.4 & -0.04 & -0.03 & -0.01 & 0.06 \\ 
        Gemini 3 Flash & -0.04 & 0.08 & 0.06 & -0.15 \\ 
        GPT mini & -0.15 & -0.19 & -0.10 & 0.02 \\ 
        Goedel v2 8b & -0.15 & -0.14 & N/A & N/A \\ 
        Gemini 3.1 Flash-Lite & -0.16 & -0.10 & -0.01 & 0.05 \\ 
        Goedel v2 32b & -0.17 & -0.10 & N/A & N/A \\ \bottomrule
    \end{tabular}
    \end{center}

    \subsection{Flowcharts for Pass@$k$ and Refine@$k$}
\begin{figure*}[h]
    \centering
    \begin{minipage}{0.42\textwidth}
        \centering
        \includegraphics[width=0.9\textwidth]{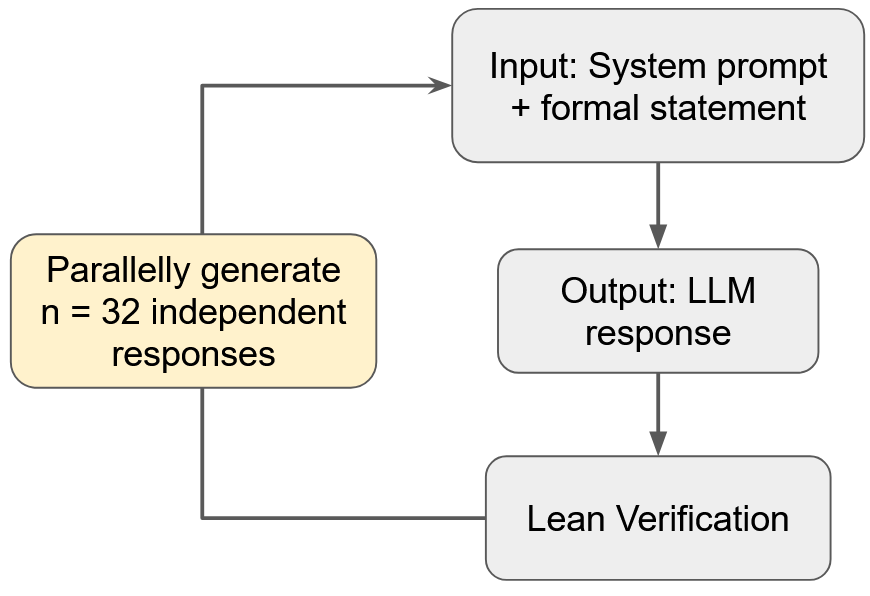} 
        \caption{Flowchart for pass@$k$}
        \label{passchart}
    \end{minipage}\hfill
    \begin{minipage}{0.57\textwidth}
        \centering
        \includegraphics[width=0.9\textwidth]{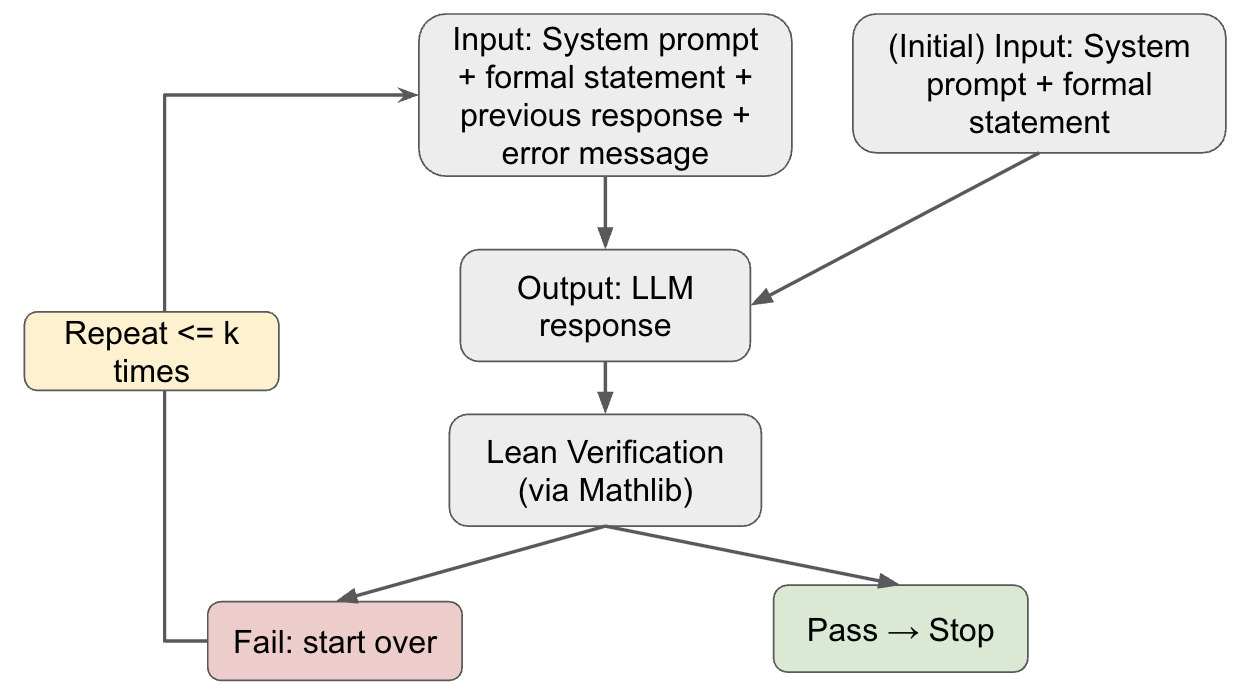} 
        \caption{Flowchart for refine@$k$}
        \label{refinechart}
    \end{minipage}
\end{figure*}

    \subsection{Prompts}
    \label{app of prompts}
    \begin{figure}[H]
    \centering
        \centering
        \includegraphics[width=0.9\textwidth]{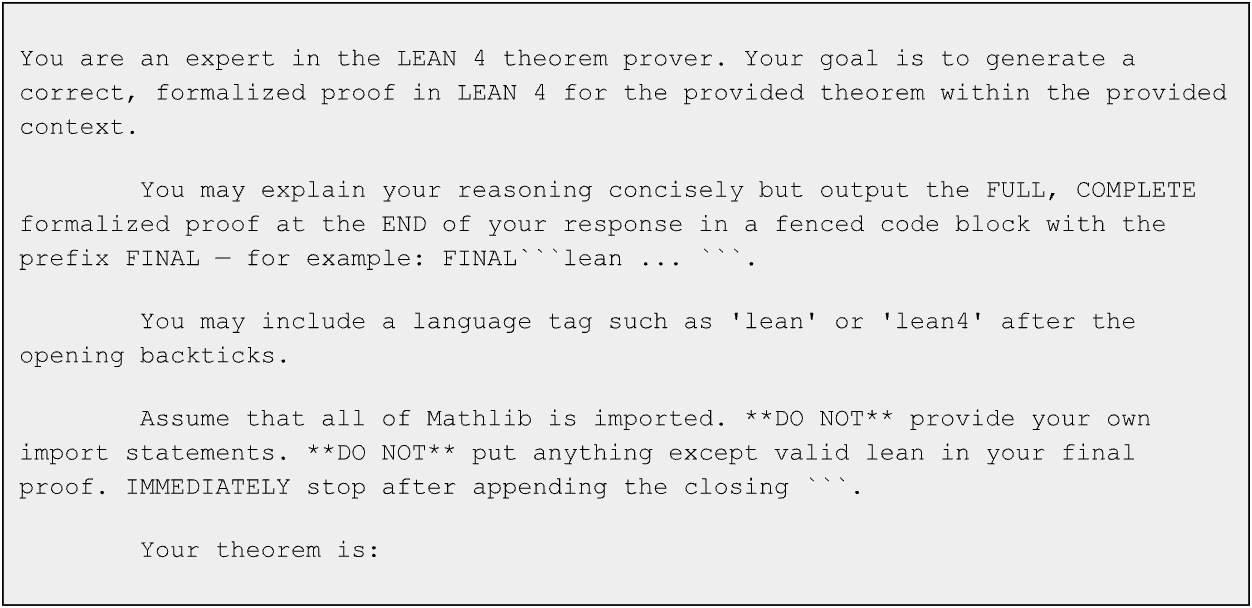} 
        \caption{Prompt for pass@$k$}
        \label{prompt1}
        \centering
        \includegraphics[width=0.9\textwidth]{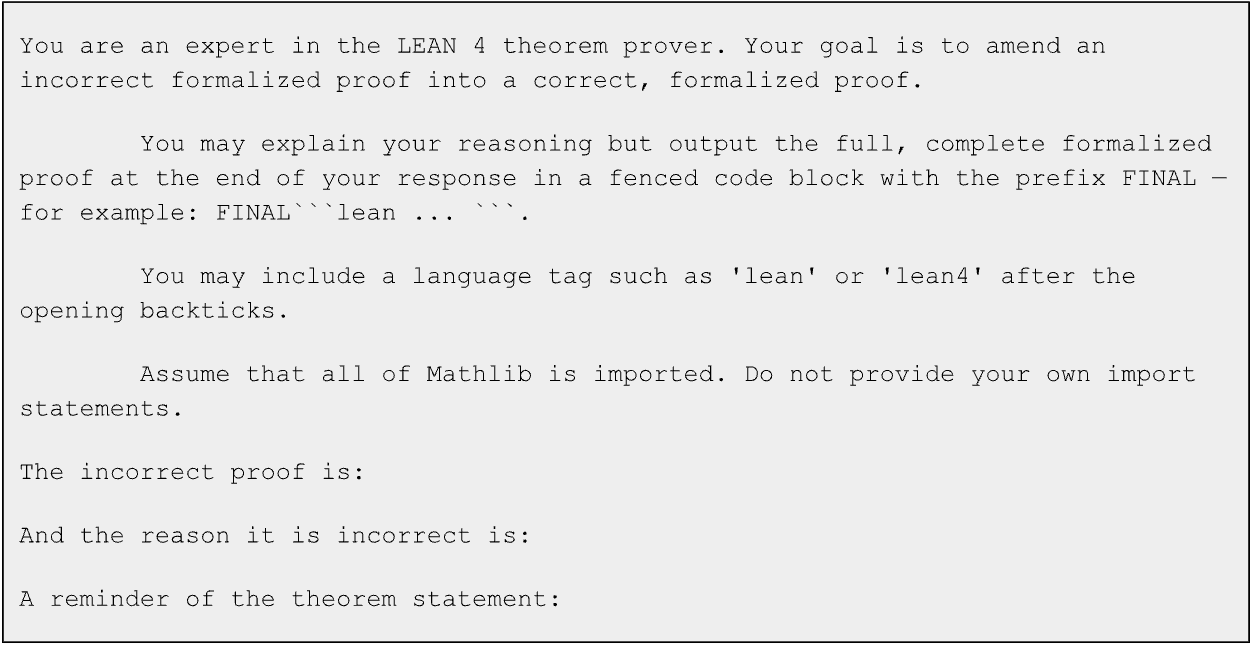} 
        \caption{Prompt for refine@$k$}
        \label{prompt2}
\end{figure}

\clearpage

\section{List of Problems used in the Datasets}
\label{appendix1}

\subsection{MiniF2F}

\begin{center}
\captionof{table}{MiniF2F problems used in subset}
\label{minif2f problems}
\begin{tabular}{|l|l|}
    \hline
    amc12b\_2020\_p21 & amc12b\_2021\_p3 \\ \hline
    amc12a\_2021\_p9 & amc12\_2000\_p6 \\ \hline
    amc12a\_2019\_p12 & amc12a\_2009\_p25 \\ \hline
    amc12a\_2021\_p25 & amc12b\_2002\_p3 \\ \hline
    amc12a\_2003\_p1 & mathd\_algebra\_419 \\ \hline
    mathd\_algebra\_33 & mathd\_algebra\_362 \\ \hline
    mathd\_algebra\_181 & mathd\_algebra\_144 \\ \hline
    mathd\_algebra\_24 & mathd\_algebra\_332 \\ \hline
    mathd\_algebra\_192 & mathd\_algebra\_412 \\ \hline
    mathd\_algebra\_270 & mathd\_algebra\_288 \\ \hline
    mathd\_algebra\_323 & algebra\_manipexpr\_apbeq2cceqiacpbceqm2 \\ \hline
    mathd\_algebra\_756 & mathd\_algebra\_156 \\ \hline
    algebra\_amgm\_sumasqdivbsqgeqsumbdiva & algebra\_manipexpr\_2erprsqpesqeqnrpnesq \\ \hline
    mathd\_algebra\_432 & aime\_1997\_p11 \\ \hline
    aime\_1983\_p1 & aime\_1988\_p8 \\ \hline
    mathd\_numbertheory\_109 & mathd\_numbertheory\_84 \\ \hline
    mathd\_numbertheory\_301 & mathd\_numbertheory\_34 \\ \hline
    mathd\_numbertheory\_690 & mathd\_numbertheory\_185 \\ \hline
    mathd\_numbertheory\_236 & mathd\_numbertheory\_342 \\ \hline
    mathd\_numbertheory\_521 & mathd\_numbertheory\_33 \\ \hline
    mathd\_numbertheory\_765 & mathd\_numbertheory\_345 \\ \hline
    mathd\_numbertheory\_237 & mathd\_numbertheory\_149 \\ \hline
    induction\_sum\_1oktkp1 & induction\_ineq\_nsqlefactn \\ \hline
    imo\_1982\_p1 & imo\_1965\_p2 \\ \hline
    imo\_2006\_p3 & imo\_1964\_p2 \\ \hline
\end{tabular}
\end{center}

\clearpage

\subsection{MiniCTX}

\begin{center}
\captionof{table}{MinCTX problems used in subset}
\label{minictx problems}
\begin{tabular}{|l|l|}
\hline
    PowerSeries.invOneSubPow\_inv\_zero\_eq\_one \\ \hline
    PadicInt.valuation\_mul \\ \hline
    List.reduceOption\_eq\_nil\_iff \\ \hline
    Condensed.hasExactLimitsOfShape \\ \hline
    ZLattice.volume\_image\_eq\_volume\_div\_covolume \\ \hline
    SetLike.covBy\_iff \\ \hline
    Polynomial.hilbertPoly\_mul\_one\_sub\_pow\_add \\ \hline
    AddValuation.ofValuation\_toValuation \\ \hline
    isBigO\_deriv\_ofReal\_cpow\_const\_atTop \\ \hline
    PowerSeries.invOneSubPow\_add \\ \hline
    HomologicalComplex.homology$\pi$\_extendHomologyIso\_hom \\ \hline
    IsCoatom.lt\_top \\ \hline
    Complex.deriv\_cpow\_const \\ \hline
    ZLattice.covolume.tendsto\_card\_le\_div'' \\ \hline
    ComplexShape.Embedding.op\_boundaryLE\_iff \\ \hline
    HomologicalComplex.extendHomologyIso\_hom\_homology$\iota$ \\ \hline
    HomologicalComplex.extendHomologyIso\_hom\_naturality \\ \hline
    PadicInt.valuation\_pow \\ \hline
    AddValuation.ofValuation\_apply \\ \hline
    Submodule.span\_range\_update\_add\_smul \\ \hline
    ZLattice.covolume.tendsto\_card\_div\_pow'' \\ \hline
    Valuation.ofAddValuation\_symm\_eq \\ \hline
    HomologicalComplex.pOpcycles\_extendOpcyclesIso\_inv \\ \hline
    CanonicallyOrderedAdd.pow\_pos \\ \hline
    PadicInt.valuation\_coe\_nonneg \\ \hline
    HomologicalComplex.homology$\pi$\_extendHomologyIso\_inv \\ \hline
    isTheta\_deriv\_ofReal\_cpow\_const\_atTop \\ \hline
    IsAtom.bot\_lt \\ \hline
    PadicInt.norm\_eq\_zpow\_neg\_valuation \\ \hline
    ComplexShape.Embedding.op\_boundaryGE\_iff \\ \hline
    Polynomial.coeff\_mul\_invOneSubPow\_eq\_hilbertPoly\_eval \\ \hline
    List.reduceOption\_eq\_concat\_iff \\ \hline
    Set.pairwiseDisjoint\_pair\_insert \\ \hline
    Polynomial.natDegree\_hilbertPoly\_of\_ne\_zero \\ \hline
    Complex.deriv\_ofReal\_cpow\_const \\ \hline
    Valuation.ofAddValuation\_apply \\ \hline
    Valuation.ofAddValuation\_toAddValuation \\ \hline
    Matrix.det\_updateCol\_add\_self \\ \hline
    DifferentiableAt.ofReal\_cpow\_const \\ \hline
    Valuation.val\_mrange\_zero \\ \hline
    Polynomial.hilbertPoly\_X\_pow\_succ \\ \hline
    StrictConvexSpace.of\_strictConvex\_unitClosedBall \\ \hline
    Subsemiring.coe\_toNonUnitalSubsemiring \\ \hline
    Matrix.det\_updateCol\_sum \\ \hline
    IsApproximateSubgroup.pow\_inter\_pow\_covBySMul\_sq\_inter\_sq \\ \hline
    CharacterModule.dual\_injective\_iff\_surjective \\ \hline
    Polynomial.hilbertPoly\_smul \\ \hline
    ZLattice.volume\_image\_eq\_volume\_div\_covolume' \\ \hline
    hasDerivAt\_ofReal\_cpow\_const \\ \hline
    HomologicalComplex.extend.leftHomologyData.lift\_d\_comp\_eq\_zero\_iff' \\ \hline
    \end{tabular}
\end{center}

\end{document}